\documentclass[letterpaper,10pt,conference]{ieeeconf}
\IEEEoverridecommandlockouts
\usepackage{booktabs}
\usepackage{times}
\usepackage{amsmath,amssymb,mathtools}
\usepackage{graphicx}
\usepackage{xcolor}
\usepackage{microtype}
\usepackage{hyperref}
\usepackage{cite}
\usepackage{float}

\graphicspath{{figures/}{./}}

\title{\LARGE \bfseries
Water Surface Swimming in a Centipede and its Robophysical Model
}

\author{%
Zhaochen J. Xu$^{1}$, Delfin Aydin$^{1}$, Abdullah Mustafa$^{1}$,
Margarita B. Levin$^{1}$, Jianfeng Lin$^{1}$,\\
Tianyu Wang$^{1}$, Daniel I. Goldman$^{1}$%
\thanks{$^{1}$All authors are with the Georgia Institute of Technology,
Atlanta, GA 30332, USA. \protect{\ttfamily\small
\{zxu699, daydan3, amustafa40,\protect\newline
mlevin32, jianf.lin, tianyuwang\}@gatech.edu,\protect\newline
daniel.goldman@physics.gatech.edu}}%
}

\begin{document}

\maketitle

\begin{abstract}
% XX THE IDEA THAT WE ARGUE THAT NON STREAMLINED LIMBED ROBOTS WILL BE USEFUL AND THUS WE NEED TO KNOW HOW TO MAKE THEM SWIM

Elongate multi-legged robots use coordinated body waves and distributed legs to move through cluttered terrestrial environments. However, as housing actuators for independent leg control can require bulky body segments, their non-streamlined body and limb structure makes it difficult to achieve swimming capability comparable to their terrestrial locomotor performance. At the water surface, we found that the multi-legged robots we tested unexpectedly moved backward: their body waves traveled in the same direction as their displacement, i.e., swimming with a direct wave. We found similar behavior in the centipede \textit{Lithobius forficatus}, which swims with a direct body wave and periodic leg movement. To study how distributed legs contribute to direct-wave swimming, we analyze animal kinematics and develop a multi-legged robophysical model that allows independent variation of leg morphology and stiffness, body-wave direction, and leg coordination. Robophysical experiments show that direct body waves produce consistent forward motion under the tested conditions and that swimming performance depends on body--leg coordination. Additionally, directionally compliant legs increase displacement from approximately 0.08 to 0.21 body lengths per cycle relative to rigid legs under matched anti-phase actuation. These findings clarify how distributed appendages contribute to surface swimming and establish gait and morphology design principles for extending multi-legged field robots from terrestrial locomotion into aquatic environments.

\end{abstract}

% xx DAN'S COMMENTS 09/04 :
% - EXPLAIN WHY LEG
% - adding legs will be a better swimmer?? or using robot to argue animal's cool.
% - why legs are important and how robophysical model validate blah

\section{Introduction}

Elongate multi-legged robots achieve effective terrestrial locomotion by coordinating waves of body undulation with distributed leg motion \cite{chong2022general,chong2023self}. This coordination exploits redundancy across many body and limb degrees of freedom, organizing a high-dimensional locomotor system through a small set of gait parameters while maintaining robust interactions with complex terrain. These capabilities make them promising platforms for field operation in cluttered environments. However, how the same body morphology can support locomotion in water remains poorly understood, limiting the basis for extending their terrestrial capabilities to aquatic environments.

% XX THIS IS AN IMPORTANT NEW BIT AND SHOULD BE IN ABSTRACT!!! (HARD TO MAKE SUCH ROBOTS STREAMLINED HUH?!?)
For example, SCUTL (Ground Control Robotics, Inc.) is a terrestrial multi-legged platform with rotary legs designed for ground contact \cite{groundcontrolrobotics_scutl}. In our exploratory water-surface trial, it moved backward relative to its body heading under a forward terrestrial command for lateral body undulation. Its body wave traveled in the same direction as its displacement, a relationship termed direct-wave swimming. Taylor's analysis of an undulating cylinder immersed in bulk fluid \cite{taylor1952analysis} predicted that sufficiently strong directional surface roughness can reverse the swimming direction of an undulating body, allowing it to move in the same direction as its bending waves. For such multi-legged robots, housing servo actuators for independent leg control often requires bulky body segments and constrains body streamlining. These bulky body segments are compatible with terrestrial locomotion. However, together with protruding legs, this non-streamlined body may alter the swimming response relative to that of smooth, slender undulatory swimmers. This observation motivates investigating how body deformation, leg coordination, and appendage morphology can enable and improve direct-wave swimming in multi-legged robots.

\begin{figure}
    \centering
    \includegraphics[width=\columnwidth]{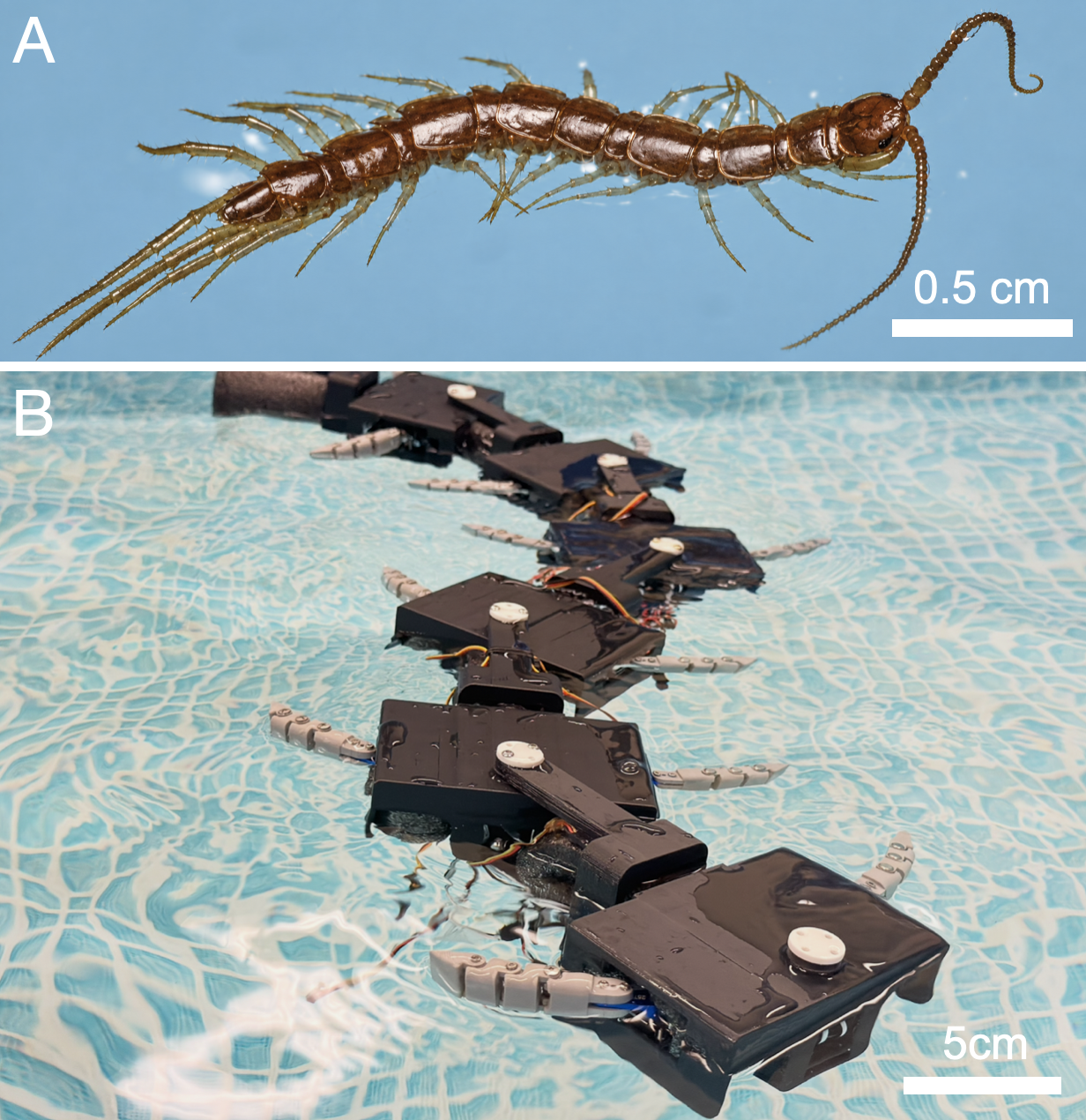}
    \caption{\textbf{Centipede surface swimming and a multi-legged robophysical model.} (A) \textit{Lithobius forficatus} swimming at the air-water interface with lateral body undulation. (B) Six-link robophysical model with independently actuated body joints and directionally compliant legs.}
    \label{fig:swimming_snapshots}
\end{figure}

Biological locomotion provides a useful basis for addressing this question. Although many fishes and eels swim using retrograde body waves that propagate opposite to the direction of locomotion \cite{santo2021convergence,tytell2004hydrodynamics}, direct waves also occur in some locomotor strategies. Terrestrial gastropods such as snails and slugs crawl using pedal waves that propagate in the direction of locomotion \cite{lai2010mechanics}. Distributed appendages introduce an additional mode of coordination. Krill, for example, generate propulsion through metachronal paddling, in which neighboring swimming appendages execute strokes with a phase lag \cite{alben2010coordination}. Swimming polychaetes combine these two features: metachronal paddling of the parapodia accompanies a direct body wave that propagates in the direction of swimming \cite{daniels2021metachronal,gemmell2024hydromechanical}. At the air-water interface, appendage mechanics can play an especially important role because propulsion depends on interactions with the free surface; water striders, for example, generate thrust by sculling specialized hydrophobic legs \cite{hu2003hydrodynamics}. Centipedes provide a particularly relevant comparison because they combine an elongate, segmented, undulating body with numerous distributed legs \cite{aoi2013instability}. Their swimming gaits also vary substantially. Submerged \textit{Scolopendra subspinipes mutilans} swims with a posteriorly propagating body wave while folding most legs against the body \cite{yasui2019decoding}, whereas surface-swimming \textit{Scolopocryptops rubiginosus} combines body undulation with anteriorly traveling leg waves \cite{kuroda2022gait}. In our observations, \textit{Lithobius forficatus} swims at the water surface with a direct body wave while continuing to oscillate its legs. Understanding how these distributed leg motions interact with body undulation to generate propulsion can reveal which features of this gait are useful for robotic surface swimming.

Robotic swimmers have employed a wide range of locomotor strategies, including body undulation, appendage-driven swimming, and amphibious gait transitions, many inspired by biological systems. Fish- and eel-inspired robots have demonstrated swimming through body and tail undulation \cite{katzschmann2018exploration,nguyen2022anguilliform}, while amphibious salamander robots have reproduced transitions between legged walking and axial swimming \cite{ijspeert2007from}. More recently, amphibious multi-legged and centipede-type robots have demonstrated locomotion across terrestrial and aquatic environments using shared body and leg structures \cite{tieng2025development,tsunoda2026development}. Robophysical studies of metachronal paddling show that inter-appendage phase affects flow structure and swimming performance \cite{ford2019hydrodynamics,ford2021role}. At the water surface, robots employ rowing \cite{sitti2007miniature}, surface-wave generation \cite{Rhee2022SurferBot}, and traveling waves along compliant fins \cite{hartmann2025flat}. Single-actuator wave (SAW) mechanisms have also enabled swimming through sagittal body undulation, with pitching body links generating thrust \cite{guetta2023novel}. Experiments with compliant swimming appendages and flexible paddles demonstrate that passive deformation can improve propulsion \cite{KWAK2017260,herreraamaya2024propulsive}. Compliant bodies and passively adaptive structures have also enabled swimming through confined and cluttered aquatic environments \cite{liu2023nonbiomorphic,wang2025aquamilr,fernandez2025aquamilr}. However, these studies do not establish how distributed legs contribute to direct-wave swimming in a laterally undulating robot at the water surface. Body bending shifts each leg's attachment point, while leg actuation and passive deformation change its motion relative to the water. Identifying effective swimming gaits therefore requires testing body-leg coordination and leg mechanics together.

% XX WRT COORDINATION OF BODY AND LIMBS.

% XX FROM HERE IS TOO MUCH DETAIL IN INTRO, PUT IN TEXT DISCUSSION WHEN INTROING ANIMAL

In this paper, we combine animal kinematics and robophysical experiments to study how body--leg coordination affects surface swimming in multi-legged robots. We use the animal's body waves and bilateral leg timing to guide controlled tests of robot gait and leg design. Our contributions are: (1) quantifying direct body and leg waves and near-anti-phase bilateral coordination in \textit{L. forficatus}; (2) using a six-link robot to show that direct body waves produce forward motion under the tested conditions and that the best-performing bilateral leg phase depends on body-wave spatial frequency; and (3) developing directionally compliant legs that increase displacement from approximately 0.08 to 0.21 body lengths per cycle relative to rigid legs under matched anti-phase actuation. We also demonstrate locomotion through obstacle arrays. These results identify gait and leg-design principles for extending terrestrial multi-legged robots into aquatic environments.

\section{Centipede Swimming}

% XX CITE PREPRINT OR IN PREP? OR SHOW CALC OF COASTING NUMBER FOR ANIMAL?

Previous measurements of surface-swimming \textit{L. forficatus} reported a coasting displacement of only \(0.05\pm0.02\) body lengths after body undulation ceased \cite{diaz2022noninertial}. This limited coasting is consistent with locomotion in a strongly dissipative regime. Here, we quantify body and leg kinematics to identify coordination patterns that can guide the robot experiments.

\subsection{Direct Body-wave Propagation in Centipede Swimming}
\label{subsec:animal_body_wave}

% \begin{figure}[t]
%     \centering
%     \includegraphics[width=\columnwidth]{figures/fig_new4_scutl.png}
%     \caption{\textbf{SCUTL direct-wave swimming} (A) Representative configurations showing surface motion with the body-wave traveling in the direction of displacement. (B) Displacement versus gait cycles with the leg wave enabled and disabled.}
%     \label{fig:scutl_leg_comparison}
% \end{figure}

Let \(s\in[0,L]\) denote arc length from head to tail and let
\(\psi(s,t)\) be the tangent orientation of the tracked centerline. The signed
centerline curvature is
\begin{equation}
    \kappa(s,t)=\frac{\partial\psi(s,t)}{\partial s}.
    \label{eq:curvature}
\end{equation}
After subtracting the time-averaged curvature, we project the field onto its
first two spatial modes \(v_1(s)\) and \(v_2(s)\),
\begin{equation}
    a_k(t)=\int_0^L
    [\kappa(s,t)-\overline{\kappa}(s)]v_k(s)\,ds,
    \qquad k\in\{1,2\},
    \label{eq:body_modes}
\end{equation}
and define the cyclic body-shape phase as
\begin{equation}
    \phi_b(t)=\operatorname{atan2}\!\left(a_2(t),a_1(t)\right).
    \label{eq:body_phase}
\end{equation}
This modal representation separates progression around the gait cycle from
the instantaneous position of the animal.
\begin{figure}[t]
    \centering
    \includegraphics[width=8.5cm]{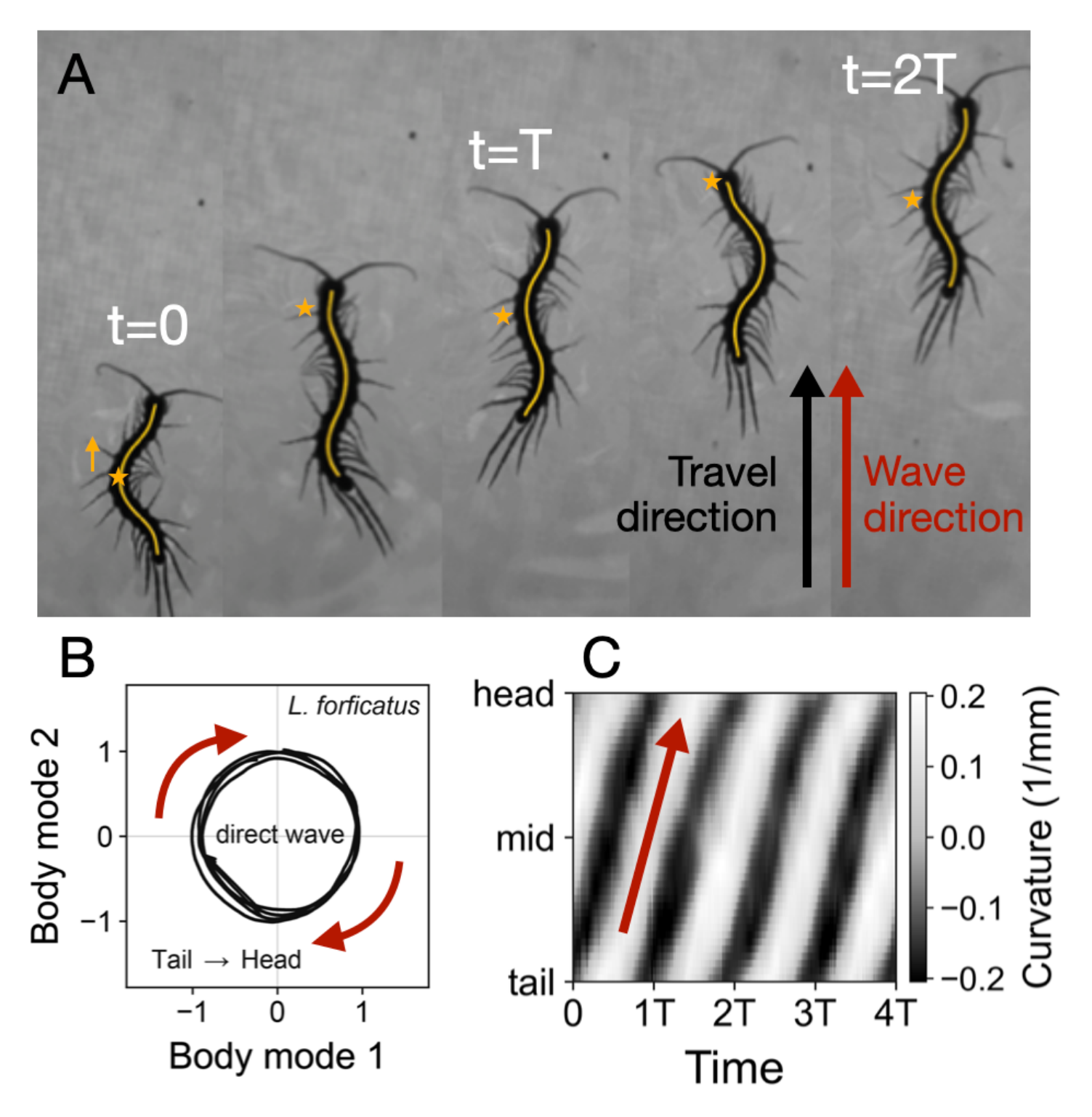}
    \caption{\textbf{\textit{L. forficatus} direct body-wave propagation during swimming.} (A) Tracked configurations between 0 and 2 body-wave periods, showing tail-to-head wave. (B) PCA profile of first two modes from Eq.~\eqref{eq:body_modes}; clockwise rotation denotes tail-to-head phase progression. (C) Centerline-curvature heatmap over four periods. Diagonal bands and the red arrow show the wave traveling from tail to head.}
    \label{fig:centipede_direct_wave}
\end{figure}

\begin{figure}[t]
    \centering
    \includegraphics[width=7.3cm]{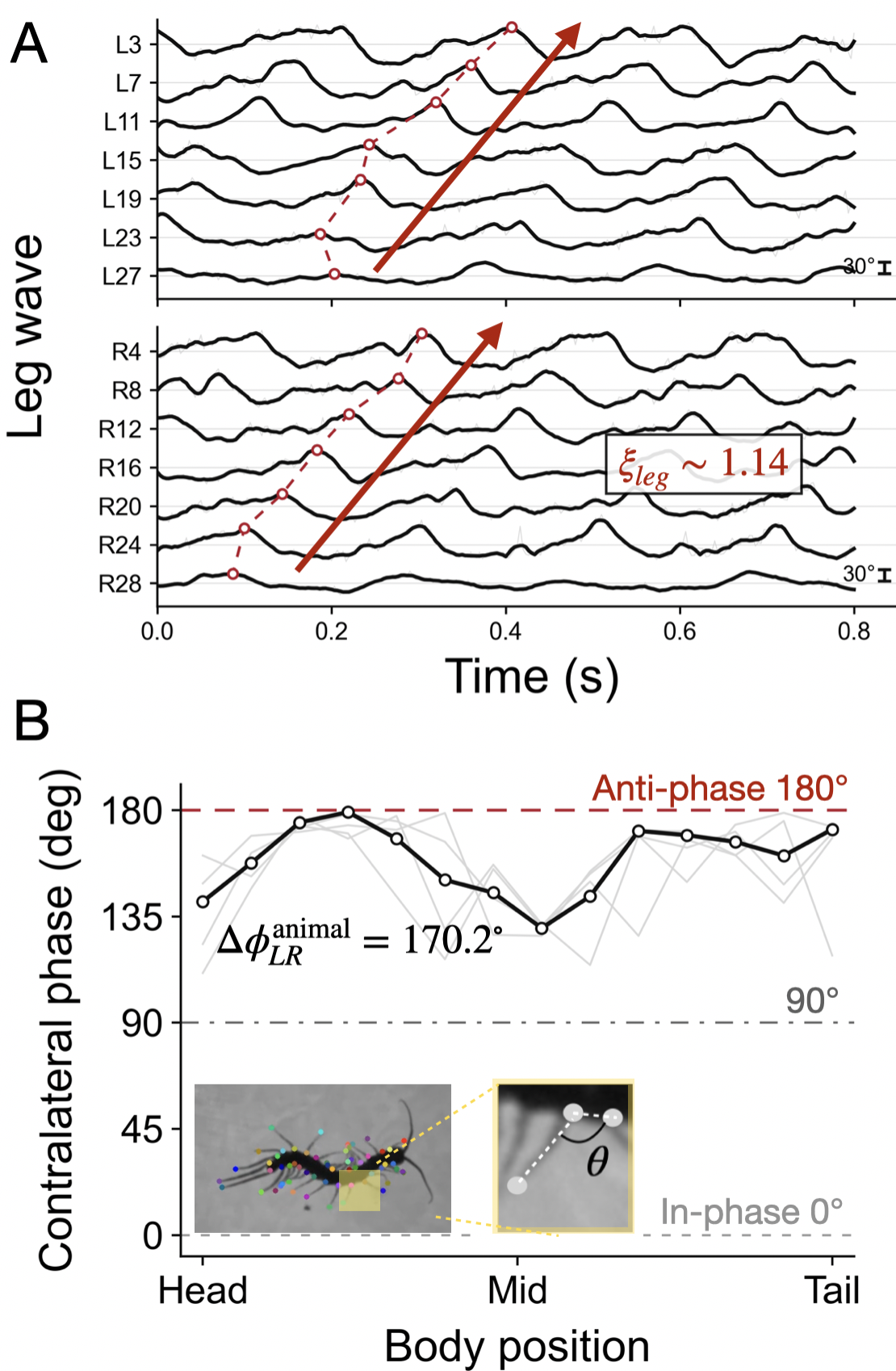}
    \caption{\textbf{\textit{L. forficatus} leg wave and contralateral coordination.} (A) Stacked leg-angle traces from the left and right sides. Successive peaks propagate from posterior legs toward anterior legs, demonstrating a direct leg wave traveling from tail to head (\(\xi_{\mathrm{leg}}\approx1.14\)). Scale bars indicate \(30^\circ\). (B) Contralateral phase between homologous left–right leg pairs along the body. The mean phase difference was \(170.2^\circ\), close to \(180^\circ\) anti-phase and distinct from \(90^\circ\) quadrature and \(0^\circ\) in-phase coordination. Gray lines show individual trial data and the black line shows the mean.}
    \label{fig:limb_wave_coordination}
\end{figure}
For the representative \textit{L. forficatus} sequence in
Fig.~\ref{fig:centipede_direct_wave}, a curvature trough advances from the tail
toward the head. The fitted phase changes by \(351.2^\circ\) from head to tail
and is nearly linear in body position (\(R^2=0.994\)). The dominant frequency
is \(4.99~\mathrm{Hz}\), corresponding to a \(0.200~\mathrm{s}\) period for the
\(23.0~\mathrm{mm}\) specimen. The clockwise orbit in the
\((a_1,a_2)\) plane provides an equivalent low-dimensional signature of this
direct wave. Here, direct denotes a tail-to-head body-wave that accompanies
forward tail-to-head swimming.

\subsection{Traveling Limb Waves}
\label{subsec:animal_limb_wave}

Leg orientation was measured relative to the local body tangent and converted
to a cyclic phase for each leg. If adjacent legs on one side have mean
unwrapped phase lag \(\Delta\phi_\ell\), then the number of waves spanning an
array of \(N_\ell\) legs is
\begin{equation}
    \xi_{\mathrm{leg}}=
    \frac{(N_\ell-1)\Delta\phi_\ell}{2\pi},
    \label{eq:leg_spatial_frequency}
\end{equation}
where phase is expressed in radians. Positive \(\Delta\phi_\ell\) denotes the
tail-to-head convention used below.

The analyzed animal sequence contains 14 tracked legs on each side over
\(0.8~\mathrm{s}\). The leg oscillation frequency is \(4.98~\mathrm{Hz}\),
close to the body-wave frequency. Linear fits to unwrapped phase give
\(30.3^\circ\) per neighboring leg on the left (\(R^2=0.955\)) and
\(32.9^\circ\) on the right (\(R^2=0.979\)). Their mean,
\(31.6^\circ\), corresponds through \eqref{eq:leg_spatial_frequency} to
\(\xi_{\mathrm{leg}}=1.14\) waves across a 14-leg side. Regional phases progress from
approximately \(0^\circ\) anteriorly to \(151^\circ\) centrally and
\(309^\circ\) posteriorly. The earlier posterior peaks therefore form a
tail-to-head traveling leg-wave.

\subsection{Contralateral Leg Coordination}
\label{subsec:animal_lr_phase}

To quantify coordination between the two sides of the animal, we compared the timing of corresponding peaks in the left and right leg-angle traces. For leg pair \(i\), \(\Delta t_i\) denotes the mean time delay between the left and right peaks, and \(T_i\) denotes the corresponding gait period. The contralateral phase difference is
\begin{equation}
\Delta\phi_{LR,i}
=
2\pi\frac{\Delta t_i}{T_i}.
\label{eq:bilateral_phase}
\end{equation}
Here, \(\Delta\phi_{LR,i}=0\) represents in-phase motion, \(\Delta\phi_{LR,i}=\pi/2\) represents quadrature, and \(\Delta\phi_{LR,i}=\pi\) represents anti-phase motion.
The mean contralateral phase across the \(N_p\) tracked leg pairs is
\begin{equation}
\overline{\Delta\phi}_{LR}
=
\frac{1}{N_p}
\sum_{i=1}^{N_p}
\Delta\phi_{LR,i}.
\label{eq:mean_bilateral_phase}
\end{equation}
The animal has a mean contralateral phase of \(\overline{\Delta\phi}_{LR}=2.97~\mathrm{rad}=170.2^\circ\) (Fig.~\ref{fig:limb_wave_coordination}B). Thus, the left and right legs move close to anti-phase, with a difference of only \(0.17~\mathrm{rad}=9.8^\circ\) from perfect anti-phase. We therefore use \(\pi~\mathrm{rad}\) as the robot anti-phase condition and compare it with \(\pi/2~\mathrm{rad}\) quadrature and \(0~\mathrm{rad}\) in-phase coordination.
\section{Robophysical Model and Gait Setup}
\label{sec:robot}

\subsection{6-link Robophysics Platform}
\label{subsec:robot_platform}

\begin{figure}[t]
    \centering
    \includegraphics[width=8cm]{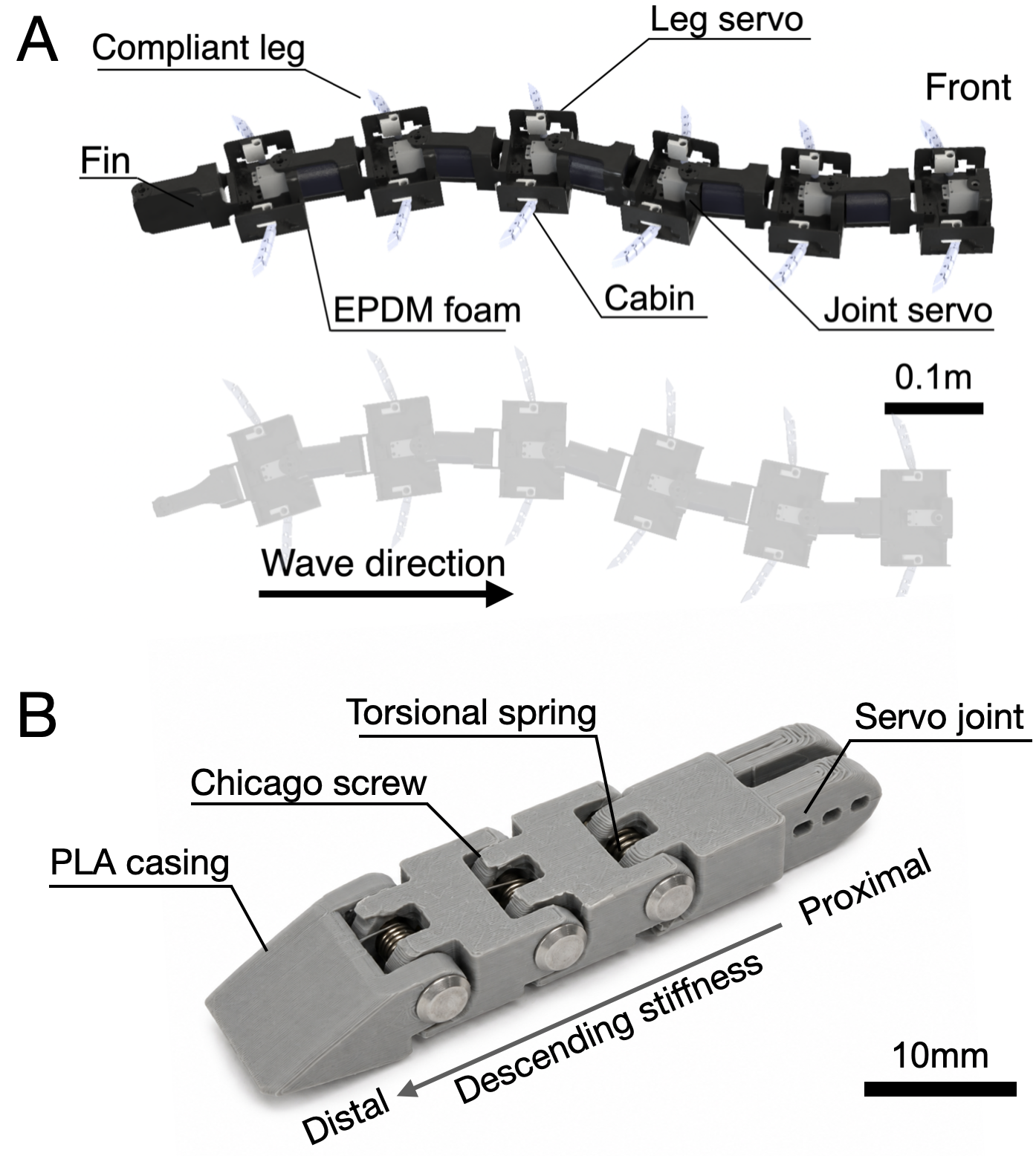}
    \caption{\textbf{Robophysical model and compliant-leg construction.} (A) Six-link robot with five actuated yaw joints and one servo-driven leg on each side of every cabin. (B) Four-link leg with stiffness decreasing toward the distal end, a servo-driven root, and three passive torsional-spring joints.}
    \label{fig:robot_design}
\end{figure}

The robophysical model consists of six rigid links, or cabins, joined in series by five actuated yaw joints that generate body undulation (Fig.~\ref{fig:robot_design}A). Each cabin is an identical, self-contained module, so link count and leg configuration can be changed independently. Each cabin carries one actuated leg per side, for twelve legs distributed along the body. Waterproof-rated servos were selected for both the body and leg joints to simplify waterproofing, limiting the required sealing to the wire splices connecting the servo leads to the ESP32 and power electronics housed off-board. Buoyancy is provided by foam sections attached along the body rather than integrated into the cabin geometry, keeping the robot reliably at the water surface.

All structural components including cabin housings, joint brackets, servo mounts, and leg links, are FDM-printed using PLA. Each cabin is printed as an integrated structure incorporating the servo housings, mounting holes, and joint interfaces that connect adjacent cabins through the body-servo horns. Each cabin measures 12.0 cm wide and 8.2 cm long. The assembled six-link robot has a total body length of 102 cm.

Each of the five body yaw joints is driven directly by its own coreless 45~kg$\cdot$cm servo with $180^\circ$ of travel. Its output shaft is coupled straight to the joint, directly setting the commanded angle $\alpha_i$ (Eq.~\eqref{eq:serpenoid}). Each leg is actuated independently at its root by a smaller coreless digital K-5 servo housed inside its cabin. An ESP32 computes the joint targets from the body and leg gait equations and sends them to a servo-driver board, which generates the corresponding PWM signals. Joint-specific offsets compensate for differences in mechanical alignment.

\begin{table}[H]
\centering
\renewcommand{\arraystretch}{1.15}
\begin{tabular}{@{}ll@{}}
\toprule
\toprule
\textbf{Mass} & Single Link: 0.237 kg \\
              & Full 6-Link Robot: 1.422 kg \\
\midrule
\textbf{Dimensions} & 12.0 cm Width \\
                    & 102 cm Length (full body) \\
\midrule
\textbf{Power} & 5.6 V, 1 A (normal operation) \\
\midrule
\textbf{Communication} & UART serial (Controller $\rightarrow$ Servo Driver) \\
\midrule
\textbf{Actuation} & Body Servos: 4.41 N$\cdot$m (stall)\\
                   & Leg Servos: 0.39 N$\cdot$m (stall) \\
\bottomrule
\bottomrule
\end{tabular}
\end{table}
\subsection{Directionally Compliant Limb Design}
\label{subsec:compliant_limbs}
Each robot cabin carries one independently actuated leg on either side of the
body. As shown in Fig.~\ref{fig:robot_design}B, each leg consists of four rigid
PLA links connected in series. The proximal link is driven directly by a
body-mounted servo, which prescribes the sweeping angle of the entire leg.
Importantly, the body-to-leg connection is an actuated joint rather than a
spring joint. The remaining three interlink joints are passive revolute joints
assembled using Chicago screws as axles and torsional springs housed within the
printed link casings.

% XX WHY DO WE NEED TWO LARGE BODY SHAPE DIAGRAMS...JUST HAVE ONE WITH ARROWS IN EITHER ROTATION
% XX WHAT ARE THESE LABELS? ALSO, IMAGES OF ROBOTS ARE WASHED OUT

\begin{figure*}[t]
    \centering
    \includegraphics[width=\textwidth]{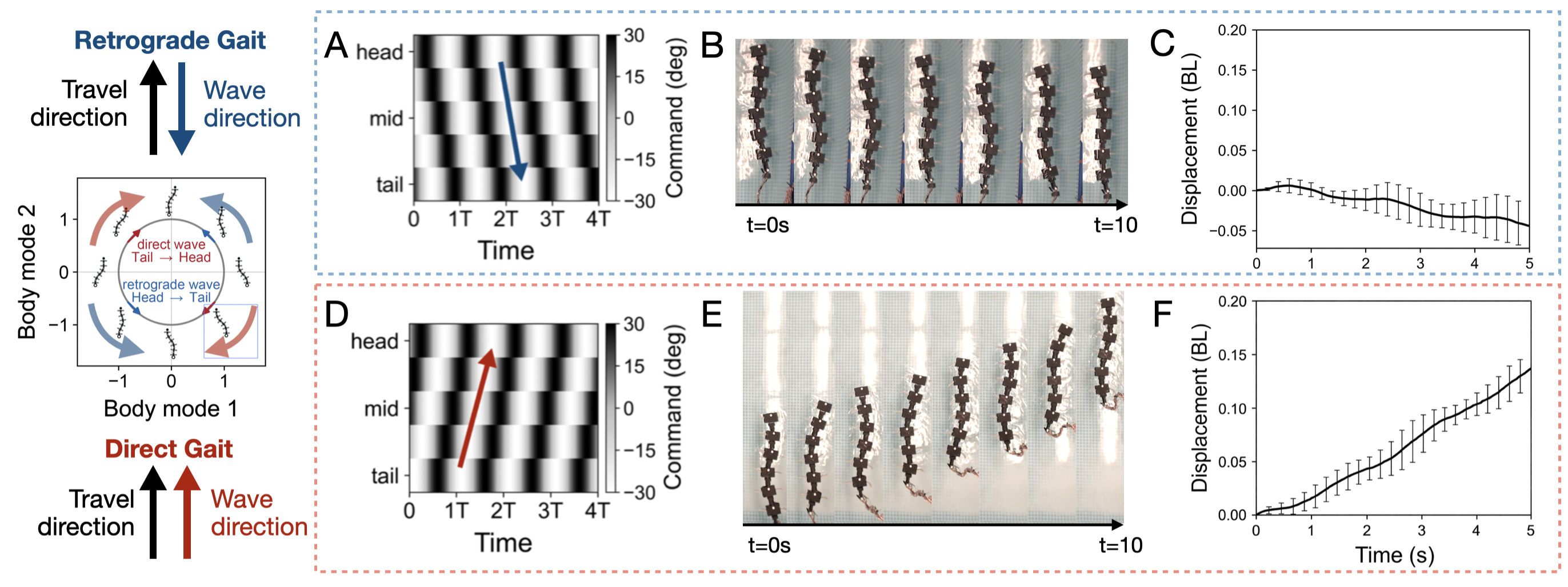}
    \caption{\textbf{Body-wave propagation and swimming direction.}
Left: body heading, wave directions, and shared body-mode diagram.
(A,D) Commanded joint-angle heatmaps.
(B,E) Representative configurations over 10~s.
(C,F) Signed displacement over 5~s, normalized by body length.
Head-to-tail waves produce backward motion, while tail-to-head
waves produce forward motion; both displacements follow same gait. Curves and error bars show means and standard deviations across three trials.}
    \label{fig:robot_wave_direction}
\end{figure*}

The passive joints use a proximal-to-distal stiffness gradient, $k_1>k_2>k_3$, where $k_i$ denotes the torsional stiffness of passive joint $i$. This distribution supports the fluid load near the leg root while allowing progressively greater reconfiguration toward the distal end. Consequently, the leg bends as a distributed structure rather than rotating primarily about a single hinge. The torsional springs also restore the leg toward its extended configuration when the external load decreases or reverses. A similar stiffness gradient in a tactile robotic appendage has been shown to provide passive reconfiguration and mechanical robustness under external loading \cite{xu2026robust}.

In the robot, one-sided mechanical stops restrict passive-joint rotation during the power stroke, maintaining an extended leg with a large projected area. During recovery, the stops disengage and the torsional springs allow the leg to bend, reducing its projected area and hydrodynamic resistance (Fig.~\ref{fig:directional_compliance}).

\subsection{Body-wave Kinematics}
\label{subsec:body_wave_control}

The robophysical platform reproduces the planar lateral body undulation observed in the centipede. We prescribe a serpenoid traveling wave \cite{hirose1993biologically} across the actuated yaw joints.

\begin{figure}[t]
    \centering
    \includegraphics[width=8cm]{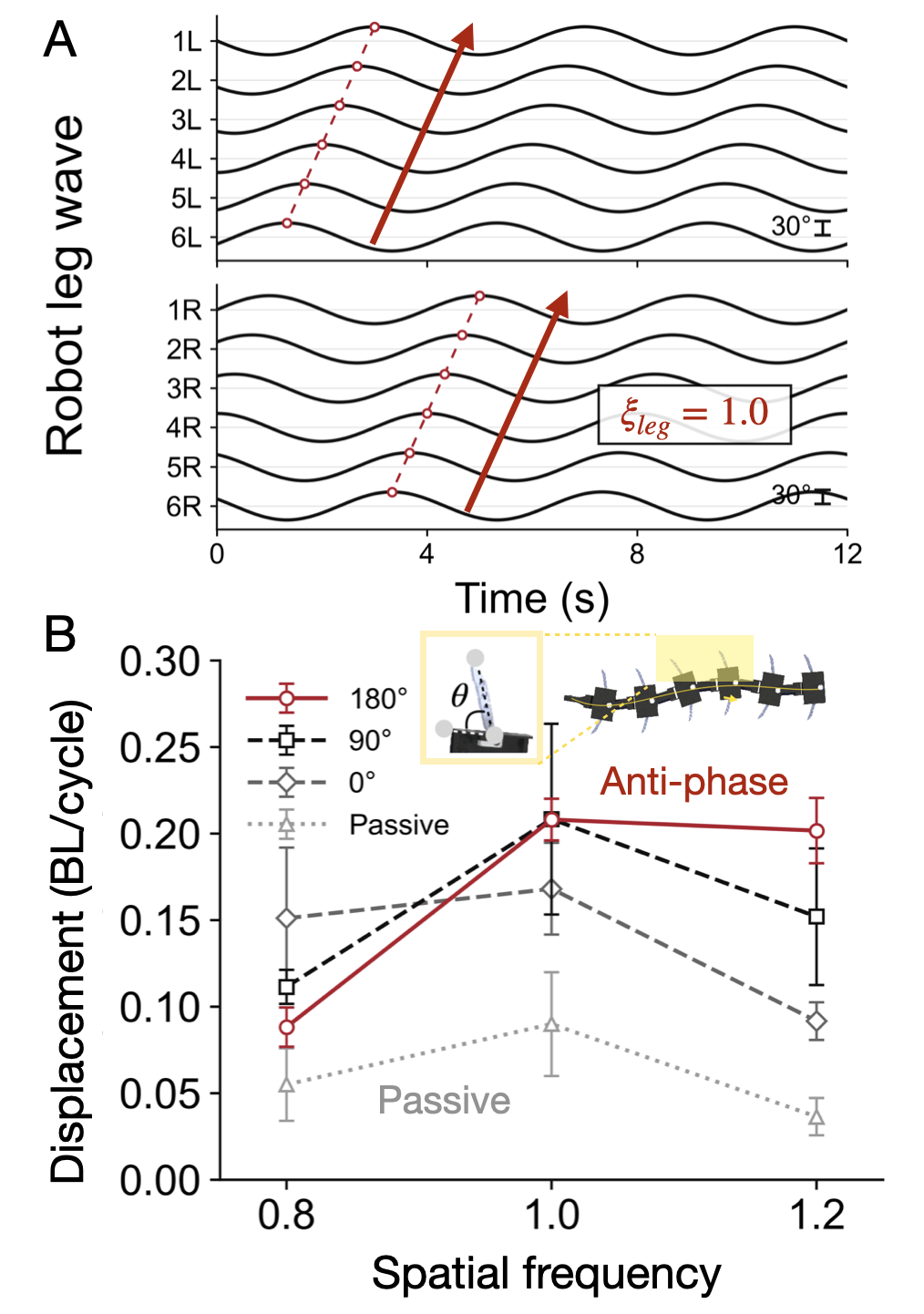}
    \caption{\textbf{Leg coordination changes direct-wave swimming.} (A) An illustrative traveling-wave command at $\xi_{\mathrm{leg}}=1.0$ produces a tail-to-head leg wave. (B) Displacement per cycle versus body-wave spatial frequency for physical anti-phase ($180^\circ$), quadrature ($90^\circ$), in-phase ($0^\circ$), and passive-leg conditions. The active-leg experiments in (B) use $\xi_{\mathrm{leg}}=0$, with no longitudinal leg phase gradient. Points and error bars are trial means and sample standard deviations. The inset defines the measured leg angle.}
    \label{fig:limb_phasing}
\end{figure}
We prescribe a traveling-wave body undulation for the $i$-th yaw joint angle $\alpha_i$ at time $t$ as
\begin{equation}
\alpha_i(t)=A_{\mathrm{body}}\,\sin\!\left(
2\pi\,\xi_b\,\frac{i}{N_i-1}-2\pi\,f\,t
\right)+\varphi,
\label{eq:serpenoid}
\end{equation}
where $A_{\mathrm{body}}$ is the amplitude, $\xi_b$ is the spatial frequency (waves along the body), $f$ is the temporal frequency, $\varphi$ is an angle offset, $i$ is the joint index, and $N_i$ is the total number of yaw joints.

\subsection{Leg-Wave Kinematics}
\label{subsec:leg_wave_control}
The sweeping leg angle at the \(i\)-th leg pair is prescribed as
\begin{equation}
\theta_i(t)=A_{\mathrm{leg}}\sin\!\left(
2\pi\xi_{\mathrm{leg}}\frac{i}{N_{\mathrm{leg}}-1}
-2\pi f_{\mathrm{leg}}t
\right)+\theta_0,
\label{eq:leg_wave}
\end{equation}
where \(A_{\mathrm{leg}}\), \(\xi_{\mathrm{leg}}\), and \(f_{\mathrm{leg}}\) are the leg-wave amplitude, spatial frequency, and temporal frequency, respectively, \(N_{\mathrm{leg}}=6\) is the number of leg-pair locations along the body, and \(\theta_0\) is the neutral leg angle. The phase difference between paired legs is \(\Delta\phi_{LR}=\phi_R-\phi_L\); for example, \(\Delta\phi_{LR}=\pi\) produces anti-phase coordination.

\section{Robophysical Experiments}
\label{sec:results}

Our goal was to identify gait and morphological changes that improve the swimming performance of multi-legged robots. Using the robophysical platform, we found that direct body waves establish forward motion, active leg coordination changes the resulting displacement, and directional leg compliance further increases propulsion. The benefit of leg motion depends on its bilateral phase and the body-wave spatial frequency. Finally, the selected direct-wave gait remains effective during repeated contacts with hydro-clutter.

\subsection{Direct Body-waves Produce Forward Swimming}
\label{subsec:direct_wave_results}

The body-mode orbits and command heatmaps illustrate opposite head-to-tail and tail-to-head wave propagation (Fig.~\ref{fig:robot_wave_direction}). Over 5~s, the tail-to-head body-wave command produces $0.14\pm0.01~\mathrm{BL}$ of forward displacement. The head-to-tail body-wave command produces $-0.04\pm0.03~\mathrm{BL}$ ($n=3$ per condition), indicating backward swimming in the direction of wave propagation. The two conditions differ in reported mean displacement by $0.18~\mathrm{BL}$ and have opposite signs. Both commands produce mean displacement in the direction of body-wave propagation, consistent with the direct-wave relation first observed in SCUTL and measured in the animal.

\subsection{Leg Coordination Modulates Performance}
\label{subsec:limb_wave_results}
\label{subsec:phasing_results}

Figure~\ref{fig:limb_phasing}A illustrates a tail-to-head leg-wave command at $\xi_{\mathrm{leg}}=1.0$. To evaluate bilateral coordination, we varied physical left-right phase with no longitudinal leg phase gradient ($\xi_{\mathrm{leg}}=0$) and compared performance with passive legs (Fig.~\ref{fig:limb_phasing}B). At $\xi_b=0.8$, mean displacement was $0.15\pm0.04~\mathrm{BL/cycle}$ for in-phase, $0.11\pm0.01$ for quadrature, and $0.09\pm0.01$ for anti-phase motion. All three active conditions had higher mean displacement than the passive reference of $0.06\pm0.02~\mathrm{BL/cycle}$.

At $\xi_b=1.0$, anti-phase and quadrature both reach approximately $0.21~\mathrm{BL/cycle}$, while in-phase motion gives $0.17\pm0.03~\mathrm{BL/cycle}$. At $\xi_b=1.2$, anti-phase maintains $0.20\pm0.02~\mathrm{BL/cycle}$, compared with $0.15\pm0.04$ for quadrature and $0.09\pm0.01$ for in-phase motion. These values again exceed the passive reference, $0.04\pm0.01~\mathrm{BL/cycle}$.

Across the active-leg conditions, displacement reaches a local maximum near \(\xi_b=1.0\), where anti-phase and quadrature coordination produce the greatest measured displacement. This phase-dependent performance shows that leg motion must be coordinated with the spatial structure of the body wave rather than selected independently.

\begin{figure}[t]
    \centering
    \includegraphics[width=8cm]{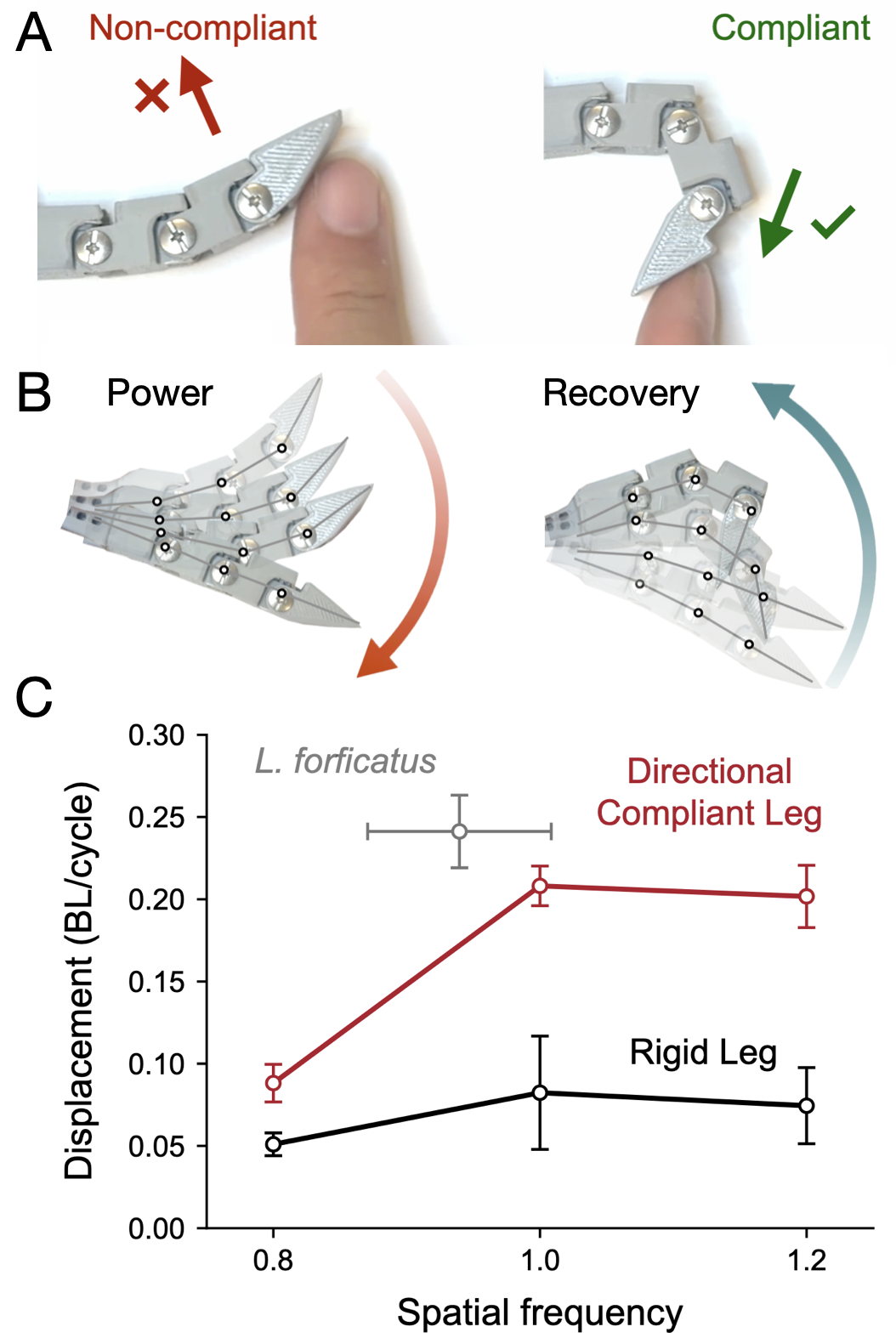}
    \caption{\textbf{Directional compliance improves displacement.} (A) One-sided stops resist bending in the power-stroke direction, while opposite loading bends the spring joints. (B) Representative leg shapes during power and recovery. (C) Displacement per cycle versus body-wave spatial frequency for directionally compliant and rigid legs at physical anti-phase. Gray shows five trials from one \textit{L. forficatus} specimen.}
    \label{fig:directional_compliance}
\end{figure}

\subsection{Directional Leg Compliance Increases Performance}
\label{subsec:compliance_results}

We designed directionally compliant legs that remain extended during the power stroke and bend during recovery. To evaluate the contribution of this passive reconfiguration to swimming performance, we compared compliant and rigid legs under matched actuation. Figure~\ref{fig:directional_compliance}A shows the one-sided joint response, and panel B shows the resulting extended and folded leg shapes under the same root actuation.

At physical anti-phase, directionally compliant legs produce more displacement than rigid legs at every tested \(\xi_b\) (Fig.~\ref{fig:directional_compliance}C). At \(\xi_b=0.8\), compliant and rigid legs produce \(0.09\pm0.01\) and \(0.05~\mathrm{BL/cycle}\), respectively. At \(\xi_b=1.0\), the corresponding values are \(0.21\pm0.01\) and \(0.08~\mathrm{BL/cycle}\); at \(\xi_b=1.2\), they are \(0.20\pm0.02\) and \(0.07~\mathrm{BL/cycle}\). Directional compliance therefore produces mean gains of approximately \(1.73\), \(2.53\), and \(2.71\) across the three body-wave spatial frequencies.

The consistent improvement supports the hypothesis that passive reconfiguration creates a more effective stroke without adding sensing or another actuator. An extended leg can maintain a larger projected area during the power stroke, while a folded leg can reduce resistance during recovery. Notably, the robot reaches a local performance optimum near $\xi_b=1.0$, close to the animal's measured spatial frequency despite the substantial difference in body size. The animal reference is $0.24\pm0.02~\mathrm{BL/cycle}$ at $\xi_b=0.94\pm0.07$ ($n=5$ trials from one specimen).

\begin{figure}[t]
    \centering
    \includegraphics[width=\columnwidth]{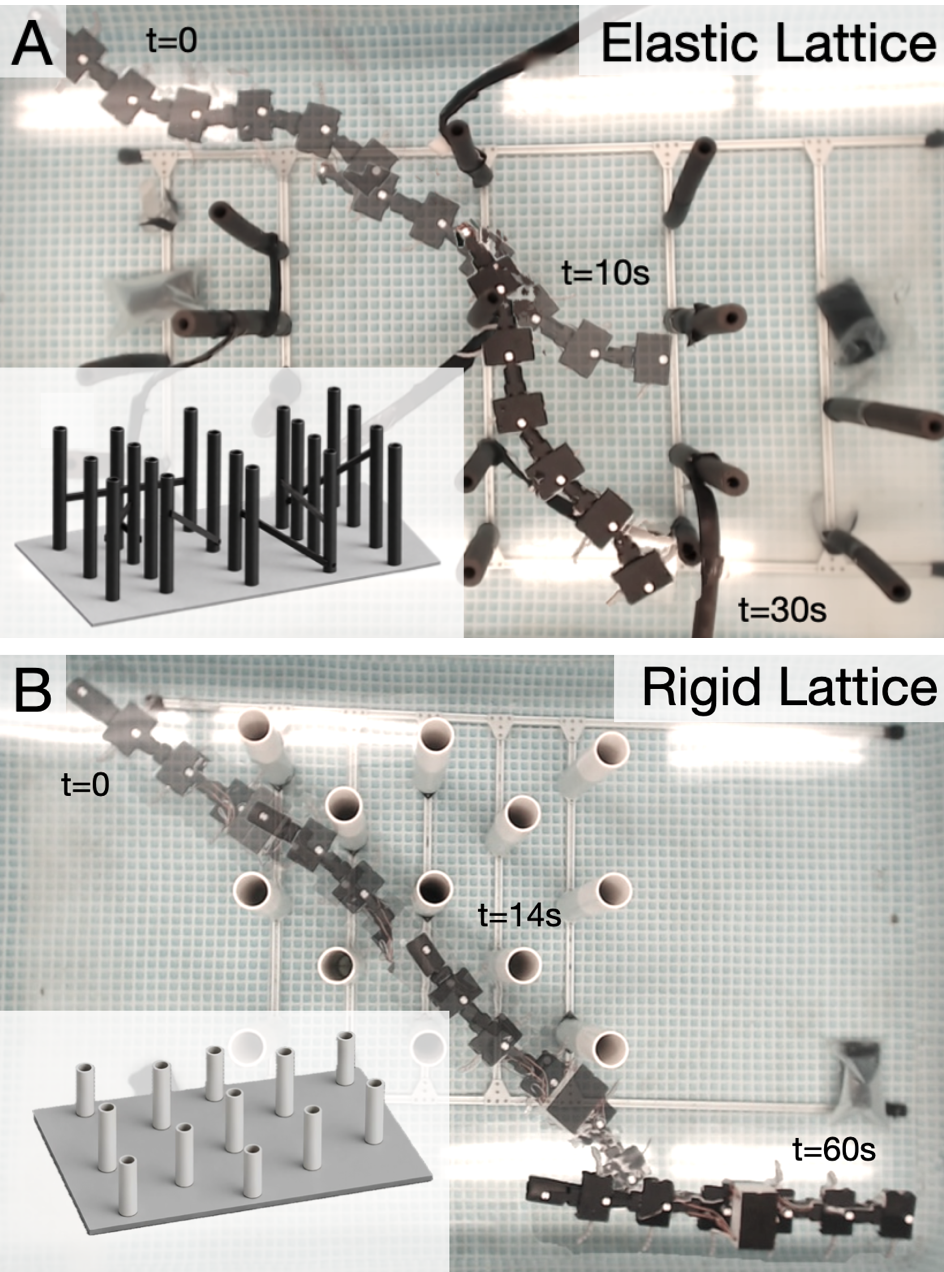}
    \caption{\textbf{Direct-wave locomotion through hydro-clutter.} Representative configurations in (A) a compliant post lattice at $t=0$, 10, and 30~s and (B) a rigid post lattice at $t=0$, 14, and 60~s. Insets show the two laboratory obstacle arrays.}
    \label{fig:hydro_clutter}
\end{figure}

\subsection{Locomotion Through Hydro-Clutter}
\label{subsec:clutter_results}

Open-water displacement does not show whether the gait remains usable when the body and legs contact obstacles. We therefore tested the robot in compliant and rigid post lattices (Fig.~\ref{fig:hydro_clutter}). In the compliant array, the robot advances through repeated contact while posts bend around the body and legs over the 30~s sequence. It also progresses through the rigid array over the 60~s sequence. These trials show that coordinated direct-wave locomotion remains effective under both fluid loading and repeated obstacle contact. The robot's passive morphological response provides a form of mechanical intelligence, allowing its appendages to reconfigure under external loads without additional sensing or control. This capability offers a practical design direction for multi-legged field robots operating in shallow water, vegetation, and other cluttered aquatic environments.

\section{Conclusion and Discussion}
\label{sec:conclusion}

Multi-legged robots provide useful terrestrial capabilities, but their distributed actuators and protruding legs constrain streamlining. Extending these platforms to water therefore requires swimming strategies compatible with their existing morphology. In this paper, we combined animal kinematics and robophysical experiments to investigate how body-wave coordination and leg mechanics enable surface swimming in multi-legged robots. Measurements of \textit{L. forficatus} identified tail-to-head body and leg waves with near-anti-phase contralateral coordination. We then used a robophysical model to examine body-wave direction and isolate the effects of leg coordination and leg morphology. Opposite body-wave commands were associated with opposite mean swimming directions, leg phase changed the resulting displacement, and directional compliance increased swimming performance. The selected gait also enabled locomotion through hydro-clutter.

The robot experiments identify two ways to improve direct-wave surface swimming: coordinating bilateral leg timing with the body wave and introducing directional leg compliance. The preferred leg phase varies with body-wave spatial frequency, while passive leg reconfiguration increases displacement under matched actuation. Notably, the robot achieved its highest measured displacement near \(\xi_b=1.0\), close to the animal's measured \(\xi_b=0.94\pm0.07\), despite the substantial difference in body size between the two systems. These findings provide specific gait and leg-design guidance for the tested multi-legged platform.

Finally, future work will investigate how the benefits of leg coordination and directional compliance depend on body size and morphology. Comparisons with other centipedes, including \textit{Scolopendra subspinipes}, which folds most legs against its body during submerged swimming \cite{yasui2019decoding}, and with polychaetes will test how these strategies extend across different body plans and swimming gaits. Establishing their range of applicability will guide the design of multi-legged robots that retain their terrestrial capabilities while gaining effective aquatic locomotion.

\section{Acknowledgment}

The authors also thank Ground Control Robotics LLC for use of the robotic platform and technical support, and Daniel Soto for his support and assistance. This work was also supported by the NSF STTR Phase I grant (2335553).

% Uncomment after creating references.bib.
% \bibliographystyle{IEEEtran}
% \bibliography{references}

\bibliographystyle{IEEEtran}
\bibliography{main}

\end{document}